\documentclass[journal]{IEEEtran}

\usepackage{graphicx}
\usepackage{amsmath,amssymb} % define this before the line numbering.
\usepackage[linesnumbered,ruled]{algorithm2e}
\usepackage{lipsum}% http://ctan.org/pkg/lipsum
\usepackage{booktabs}

\ifCLASSINFOpdf
\else
\fi
\begin{document}
\title{Iterative Spectral Clustering for Unsupervised Object Localization}

\author{\IEEEauthorblockN{Aditya Vora\IEEEauthorrefmark{1} and
Shanmuganathan Raman\IEEEauthorrefmark{2}\\}
\IEEEauthorblockA{Electrical Engineering,
Indian Institute of Technology Gandhinagar, Gujarat, India, 382355\\
Email: \IEEEauthorrefmark{1}aditya.vora@iitgn.ac.in,
\IEEEauthorrefmark{2}shanmuga@iitgn.ac.in}}


\markboth{Journal of \LaTeX\ Class Files,~Vol.~14, No.~8, August~2015}%
{Shell \MakeLowercase{\textit{et al.}}: Bare Demo of IEEEtran.cls for IEEE Journals}



\maketitle

\begin{abstract}
This paper addresses the problem of unsupervised object localization in an image. Unlike previous supervised and weakly supervised algorithms that require bounding box or image level annotations for training classifiers in order to learn features representing the object, we propose a simple yet effective technique for localization using iterative spectral clustering. This iterative spectral clustering approach along with appropriate cluster selection strategy in each iteration naturally helps in searching of object region in the image. In order to estimate the final localization window, we group the proposals obtained from the iterative spectral clustering step based on the perceptual similarity, and average the coordinates of the proposals from the top scoring groups. We benchmark our algorithm on challenging datasets like Object Discovery and PASCAL VOC 2007, achieving an average CorLoc percentage of $51\%$ and $35\%$ respectively which is comparable to various other weakly supervised algorithms despite being completely unsupervised.
\end{abstract}


\section{Introduction}
\label{intro}

Object localization is an important computer vision problem where the task is to estimate precise bounding boxes around all categories of the objects present in the given image. Due to the intra-class variations, occlusion, and background clutter present in the real-world images, this becomes a challenging problem to solve. Compared to image classification, localization involves estimating precise location of an object in the image. Therefore, it proves to be more difficult problem to solve. Object localization is useful in several image understanding tasks like separating the foreground from the background, object recognition, and segmentation. Previous fully-supervised approaches relied on sliding window search in order to search for an object in the image. Because of their inefficiency in terms of speed, several efficient sub-window search algorithms were proposed which work quite well in localizing the object in an image (\cite{lampert2009efficient}). However, these techniques require strong supervision in the form of manually-annotated bounding boxes on locations of all  the object categories in an image. Acquiring such human annotations for training accurate classifiers is a cumbersome task and is prone to human errors. As a result, supervised techniques for object localization do not prove to be useful in resource restricted settings. In order to overcome the huge manual efforts required in annotations of objects in the image in supervised learning algorithms, several weakly supervised approaches were proposed. Rather than bounding box annotations of target instances, weakly supervised learning focuses on image level labelling which is based on the presence/absence of target object instances in an image (\cite{gokberk2014multi,deselaers2010localizing,nguyen2009weakly,hoai2014learning,song2014learning}). Though these techniques work well in terms of localization accuracy, they still require human annotation efforts especially when the training data is large. 

In an effort to make the task of object localization completely unsupervised, various object co-localization algorithms were proposed which try to localize an object across multiple images (\cite{sivic2005discovering,russell2006using,grauman2006unsupervised,kim2009unsupervised,cho2015unsupervised}) without any supervision.  As co-localization algorithms assume that each image has the same target object instance that needs to be localized (\cite{grauman2006unsupervised,kim2009unsupervised}), it imports some sort of supervision to the entire localization process thus making the entire task easier to solve using techniques like  proposal matching (\cite{cho2015unsupervised}) and clustering (\cite{tang2014co}) across images. In contrast to these works, the work presented in this paper focuses on localizing a single object instance in an image in a completely unsupervised fashion. To the best of our knowledge there is no previous work that tries to solve this problem in an unsupervised way. In this work, we do not make any assumptions like that in co-localization algorithms, thus making the entire problem more practical and challenging one to be solved. Further, it is an important problem to be addressed because of the following reasons: (1) Proposed work is an unsupervised approach for object localization. As mentioned previously, all the manual labour required in annotating the data with accurate bounding box regions around the target object instances will not be required, which saves the resources as well as the training time. (2) Apart from being fully automatic and unsupervised, our technique is easy-to-fit in the current state-of-the-art object recognition pipelines like RCNN (\cite{girshick2014rich}). Thus unlike current system, we do not need to classify each of the thousands of object proposals generated from an object proposal algorithm individually. Instead, we can localize the object directly in the input test image and then provide this localized object to the CNN pipeline that will classify the object appropriately. Such a type of functionality is not available with co-localization techniques. This restricts their applicability in real-world scenarios. 

To solve this problem in an unsupervised manner, we start with extracting thousands of object proposals from the input image using an off-the-shelf object proposal algorithm (\cite{zitnick2014edge,uijlings2013selective}). We then try to filter out number of object proposals effectively in such a way that after the entire proposal filtering process, a good set of object proposals that contain the object are retained. In order to achieve this, we formulate the problem as an undirected graph problem and perform spectral clustering on the constructed graph. This will split the set of proposals that can be discriminated based on the selected feature space. However, one iteration of spectral clustering would not be enough to filter the proposals by a significant amount. As a result, we repeat the process for a number of iterations after selecting appropriate cluster for subsequent partitioning. We compute a cluster score after each iteration and select the cluster that has higher score for further partitioning in the next iteration and discard all the proposals in the cluster having lower score. After this filtering step, we then estimate the final localized window by grouping the proposals based on the perceptual similarity among the proposals. We then pick top scoring groups and take the mean of coordinates of proposals present in that groups in order to get the final localized window.

The main contributions of this paper are summarized as follows: (1) A completely unsupervised object localization algorithm for an image containing a single object is presented and benchmarked on a challenging datasets like Object Discovery (\cite{Rubinstein13Unsupervised}) and PASCAL VOC 2007 (\cite{pascal-voc-2007}). (2) An iterative spectral clustering approach along with an appropriate cluster selection strategy is proposed which naturally helps in searching of object region in the image. This entire process takes place in a completely unsupervised fashion. (3) Proposal grouping technique is proposed which helps in estimating the final localized window in the image.   

%The rest of the paper is organized as follows. Section \ref{rw} describes the related contributions in solving the problem of object localization. Detailed explanation of the entire algorithm is described in Section \ref{pa}. Section \ref{ev} describes about the result obtained by benchmarking the algorithm on various datasets. Section \ref{conc} which is the last section describes the conclusion and future work.

\begin{figure*}[t]
\begin{center}
   \includegraphics[width=0.95\linewidth]{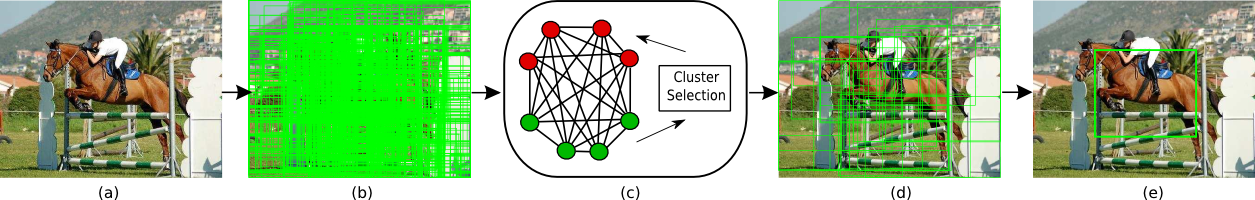}
\end{center}
\caption{\textbf{Overview of the proposed approach}. (a) Input image. (b) Extraction of object proposals ($N$). (c) Iterative spectral clustering stage. Construct a graph with nodes as HOG descriptor extracted from each proposal, red nodes $=$ foreground region and green nodes $=$ background nodes. Spectral clustering and cluster selection continues for a number of iterations with the graph being updated in each iteration based on the cluster selected. The process continues till the stopping criteria ($T$) is met. (d) Result after iterative spectral clustering after reaching stopping criteria ($T$). (e) Final estimated localized window $b_{final}$.} 
\label{Fig:pipeline}
\end{figure*}

\section{Related Work}
\label{rw}
Previous works in object localization are described below based on the decreasing order of supervision.

\textbf{Supervised approaches} for object localization involves \textit{sliding window} approaches that apply a classifier subsequently to subimages, thus obtaining a classification map. Indicator of the object region is obtained from the classification map as the region with the maximum score. As an average size image will have a lot of pixels, scanning all of them and deriving the classification map is a computationally expensive task. \cite{lampert2009efficient} and \cite{villamizar2012bootstrapping} tried to come up with a more efficient solution by proposing an \textit{efficient subwindow search} for object localization which does not suffer from the above mentioned drawbacks. This scheme helps to optimize the quality function over all the possible subregions of the image with fewer number of classification evaluations and thus making the algorithm run in linear time or faster. \cite{blaschko2009object} introduced the concept of \textit{global and local context kernels} that tries to combine different context models into a single discriminative classifier. \cite{sermanet2013overfeat} proposed an integrated framework for image classification, localization and detection. They efficiently implemented a multiscale and sliding window approach within a convolutional neural network (CNN). They treat localization as a regression problem where the final layer is involved in predicting the coordinates of the bounding box. This entire system is trained end-to-end with bounding box annotations from the ImageNet dataset (\cite{krizhevsky2012imagenet}). 

\textbf{Weakly-supervised approaches} for localization can be divided into $4$ categories: (i) exhaustive search technique (\cite{nguyen2009weakly,hoai2014learning,pandey2011scene,zhang2010weakly}), (ii) multiple instance learning (\cite{qi2007incorporating,galleguillos2008weakly,siva2012defence,cinbis2017weakly,vijayanarasimhan2008keywords,wang2014weakly}),
(iii) inter-intra-class modelling (\cite{deselaers2010localizing,deselaers2012weakly,russakovsky2012object,
siva2011weakly,wang2013weakly}), and (iv) topic model (\cite{shi2013bayesian,sivic2005discovering}). Exhaustive search techniques try to learn discriminative sub-window classifiers from the weakly labelled data and then based on the scores of the most discriminative local regions of the image, they try to estimate the final localization window. Multiple-instance learning approaches try to learn various object categories from the bag of positive and negative labelled images. Different multiple-instance learning algorithms try to exploit various aspects associated with the image. For example,  \cite{wang2014weakly} tries to model the latent categories of the image like sky and grass in order to improve the overall localization accuracy. \cite{vijayanarasimhan2008keywords} proposed an alternative learning approach where they trained robust category models from images returned by keyword-based search engines. In order to improve the quality of the object regions, along with the inter-class models, researchers also model intra-class relations to improve the similarity of the regions within the same object class.   

\textbf{Co-localization approaches:} \cite{tang2014co} proposed an image-box formulation for solving object co-localization problem, where they simultaneously localize object of the same class across a set of images. \cite{cho2015unsupervised} generalized the task of object localization by relaxing the condition that each image should contain the object from the same category. \cite{kim2009unsupervised} proposed an iterative link analysis technique in order to estimate the region of interest in the image. \cite{grauman2006unsupervised} proposed an approach for colocalization which is based on the partial correspondences and the clustering of local features.  

\section{Proposed Approach}
\label{pa}
In this section, we describe the entire pipeline for the unsupervised object localization. The summary of the entire pipeline is shown in Fig. \ref{Fig:pipeline}.

\subsection{Object proposals extraction and scoring}
We generate object proposals from the input image using off-the-shelf object proposal generation algorithm known as EdgeBoxes (\cite{zitnick2014edge}), which generates object proposals based on the edge information present in the image. We chose this technique for extracting object proposals as it is capable of generating proposals with high recall at a very fast rate. It extracts $\sim$1000 object proposals per image in around 0.25 secs achieving an object recall of over $96\%$ at an overlap of $0.5$ on the PASCAL VOC dataset (\cite{pascal-voc-2007}).

After the extraction of object proposals $B=\{b_{1},b_{2},...,b_{N}\}$ from the image $I$, we score each proposal based on the probability that the region contains an object. Here, we extract $N=1000$ object proposals. EdgeBoxes algorithm computes objectness scores $s_{obj}$ for each proposal in $B$ which is based on the fact that the number of edge contours that are wholly contained by the proposal is indicative of the proposal containing an object. We combine appearance score of each proposal with the saliency score in order to compute the overall score of each proposal. In order to do this, we compute the saliency map $S$ of the input image $I$ using the saliency algorithm proposed by \cite{margolin2013makes}. From the saliency map $S$, we compute the average saliency score for each object proposal which gives saliency score $s_{sal}$ for that particular proposal. After this, the overall score $s_{i}$ of all the object proposals in the set $B$ are computed as $s_{i}=s_{obj}\times s_{sal}$, where $i=1$ to $N$. As a result, proposals that have high objectness score and those that cover the salient region of the image will have high overall score.

\subsection{Iterative spectral clustering for proposal filtering}
\begin{figure}
\centering
\includegraphics[scale=2]{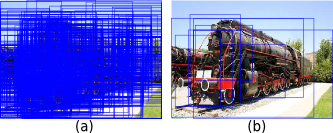}
\caption{\textbf{Results of iterative spectral clustering}. (a) Candidate regions extracted from the input image from the PASCAL VOC 2007 dataset (\cite{pascal-voc-2007}). (b) Filtered object proposals of the image.}
\label{Fig:prop_filt}
\end{figure}

Given a set of object proposals $B=\{b_{1},b_{2},...,b_{N}\}$ and proposal scores $s=\{s_{1},s_{2},...,s_{N}\}$ ($N=1000$), we need to effectively select a subset of proposals that have a high probability of containing an object. We model the feature similarity among the object proposals using an undirected graph. For each proposal $b_{i} \in B$, we extract a HOG descriptor $f_{i}$ on a $8\times8$ grid (\cite{dalal2005histograms}). We model graph $W$ based on the HOG feature similarity among the computed proposals in the set $B$. Each node of the graph is the computed HOG descriptor $f_{i}$ and each edge of the graph is weighted by $w_{ij}$, where $w_{ij}$ is the gaussian similarity score computed as $w(f_{i},f_{j}) = exp(-\frac{\|{f_{i}-f_{j}}\|^{2}}{2\sigma^{2}})$. Here the parameter $\sigma$ controls the width of the neighbourhood. For our experiments we select $\sigma$ to be $0.05\times \max(\|f_{i}-f_{j}\|)$. After constructing the graph $W$, we then compute the normalized Laplacian matrix of the graph W as $L = I-D^{\frac{-1}{2}}WD^{\frac{-1}{2}}$, where $D$ is the diagonal matrix composed of the sum of rows of $W$. This choice is motivated by the work of \cite{shi2000normalized} who showed that selecting the second smallest eigenvector of the normalized Laplacian graph $L$ leads to bi-partitioning of the graph. As HOG features are able to discriminate between the foreground and background object proposals through the modelled graph, this bi-partitioning will try to partition these proposals into two separate clusters. However, one step of spectral clustering will not be able to select highly localized proposals from the huge set of object proposals. As a result, we perform a few iterations of spectral clustering until we are left with a few highly localized object proposals from which we can estimate the final detected window. In order to select a cluster for subsequent partitioning, we compute a cluster score by taking the average of all the scores $s$ among all the proposals present in the cluster. We pick the cluster with higher score for further partitioning and discard the proposals in the cluster with lower score. We continue this iterative spectral clustering until the number of proposals are less than stopping criteria $T$ (here, $T=100$). As it can be seen in Fig. \ref{Fig:prop_filt}(b) that after iterative spectral clustering, highly localized object proposals are retained and rest of the proposals are discarded from the original set of proposals (Fig. \ref{Fig:prop_filt}(a)).    

\begin{algorithm}[h]
\caption{Pipeline for Object Localization}
\label{mainalgo}
 \KwIn{Image $I$} 
 \KwResult{Final localization window, $b_{final}$}
 \textbf{Procedure:}\\
 Set Number of proposals, N.\\
 %Set cell size for HOG features. \\
 Set stop criteria for iterative spectral clustering, $T$.\\
 Set $k$ for NN search and $C$ for number of groups to select.\\
 Extract object proposals $B=\{b_{1},b_{2},...,b_{N}\}$ from $I$.\\
 Compute saliency map $S$ of $I$.\\
 Compute saliency score $s_{sal}$ for each proposal.\\
 Compute overall score $s_{i} = s_{obj}*s_{sal}$, $i=1$ to $N$.\\
 Compute HOG features for $b_{i} \in B$, i.e., $F=\{f_{1},f_{2},...,f_{N}\}$.\\
\While{$N<=T$}{ 
 Construct similarity graph $W$ on $F$ feature set.\\
 Compute normalized laplacian $L$ and perform bi-partitioning.\\
 Compute cluster scores and select appropriate cluster. \\
 Update proposals $B$.\\ 
 }
 Compute Dense SIFT features $\hat{F}=\{\hat{f}_{1},\hat{f}_{2},...,\hat{f}_{K}\}$, for $\hat{b}_{i} \in \hat{B}$, where $\hat{B}=\{\hat{b}_{1},\hat{b}_{2},...,\hat{b}_{K}\}$, where $K<T$.\\
 \textit{k}-NN for $\hat{b}_{i} \in \hat{B}$ resulting in $G=\{g_{1},g_{2},...,g_{K}\}$ groups.\\
 Compute scores for $g_{i} \in G$ using Eq. \ref{eq1} and pick top-\textit{C} groups.\\
 Compute mean of all coordinates of proposals to get the final localized window $b_{final}$.   
 
\end{algorithm}

\subsection{Estimating the final localized window}

After the iterative spectral clustering step, the proposals obtained from the final cluster localizes the object region in the image, as it can be seen in Fig. \ref{Fig:prop_filt}(b). This is due to the reason that the proposals in the final cluster have a high HOG feature similarity and they stand out based on the cluster score during the entire clustering process. We need to estimate a tight localization window around the object from these proposals in order to get good localization accuracy in a completely unsupervised fashion. One simple way to do this is to take the mean of all the proposals obtained after the iterative spectral clustering step. However, such a naive technique is prone to outliers (those object proposals that are larger in area will have less portion of object covered and have more background region) which will affect the overall mean of all the windows and thus affect the final localization accuracy. As a result, we need to come up with a better strategy to estimate the final window location which is immune to such outliers and thus resulting in an accurate localization. 

In visual perception, a high contrast difference exists between the foreground and the background regions and low contrast difference exists between the foreground-foreground and background-background regions. We exploit this idea in order to get a good localization window. We group the object proposals obtained after clustering by considering each proposal as a seed proposal and selecting other members of the group as the proposals that have a high feature similarity with the seed proposal. We score each group based on the feature similarity of group's member proposals with the seed proposal and on the score of each proposal in the group. The main difference between iterative spectral clustering step and proposal grouping strategy is that, iterative spectral clustering helps in effectively filtering huge set of proposals at a very fast rate by reducing the number of proposals by about half in each iteration. However, each proposal remaining after the iterative spectral clustering step will have some implication on the final localization accuracy as it can be seen in Fig. \ref{Fig:prop_filt}(b). As a result, continuing iterative spectral clustering on these proposals will discard many such proposals that highly contribute towards the localization accuracy. Because of this, we come up with a proposal grouping technique that considers each proposals implication on localization accuracy through the contrast differences between the foreground and background regions, thus helping in better localization window estimation.

After the iterative spectral clustering step, we are left with a set of proposals $\hat{B}=\{\hat{b}_{1},\hat{b}_{2},...,\hat{b}_{K}\}$, where $K<T$. For each object proposal $\hat{b}_{i}\in\hat{B}$, we extract dense SIFT features as local discriminative features every $4$ pixels and vector quantize each descriptor into a $1,000$ word codebook. For each proposal, we pool the SIFT features within the proposal using $1\times1$ and $3\times3$ spatial pyramid matching (SPM) (\cite{lazebnik2006beyond}) pooling regions to generate a $d = 10,000$ dimensional feature descriptor for each box, similar to the one used by \cite{tang2014co} for co-localization. We then normalize each feature descriptor using $L_{2}$ norm. As a result of this, we get a set of features $\hat{F}=\{\hat{f}_{1},\hat{f}_{2},...,\hat{f}_{K}\}$ for each proposal in $\hat{B}$. We then group the set of object proposals that are perceptually similar. For this, we consider each proposal from the set $\hat{B}$ as the seed proposal and find the \textit{k}-nearest neighbours of this seed proposal using the $\hat{F}$ feature set. Here we select $k=10$. We do this \textit{k}-nearest neighbour search for all proposals. After this, we get $G=\{g_{1},g_{2},...,g_{K}\}$ groups of object proposals with each group containing one member as a seed proposal and remaining members of the group with proposals that have higher similarity with the seed proposal. In order to get the final localization window, we need to select best groups from this set $G$. We compute the score for each group based on the feature similarity among the proposals in that group and the proposal score of the member object proposals in that group. The score for each group $g_{i} \in G$ is given as: 
\begin{equation}
s_{group}(i) = \sum_{j=2}^{\textit{k}} s(\hat{b}_{j}) (\hat{f}_{j}^{\top}\hat{f}_{i})
\label{eq1}
\end{equation} 
Here $i$ is the seed proposal of the group and $j$ is the index of the group members and thus the range is from $2$ to $k$. We pick top-$C$ proposal groups (here, $C=5$) that have maximum group scores and obtain a final set of proposals by taking the union of all proposals in the top-$C$ groups (i.e., many proposals in the top-$C$ groups would be common, thus those proposals that are common will be considered only once in the final set if we take the union). We then take the mean of all the coordinates of the bounding boxes in the final set of proposals in order to obtain the final detection window $b_{final}$ as shown in Figure. \ref{Fig:pipeline}(e).

\begin{figure*}
\centering
\includegraphics[scale=1.7]{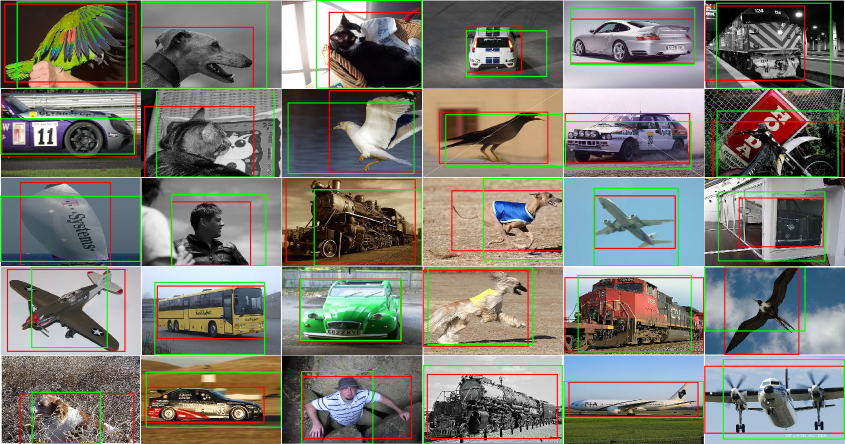}
\caption{Results of localization on PASCAL VOC dataset (\cite{pascal-voc-2007}). Green box $=$ Estimated Window, Red box $=$ Ground Truth.}
\label{Fig:results}
\end{figure*}

\section{Evaluation}
\label{ev}
In order to evaluate our algorithm, we conducted experiments on two realistic datasets, the Object Discovery dataset (\cite{Rubinstein13Unsupervised}) and PASCAL VOC 2007 dataset (\cite{pascal-voc-2007}) and compare the results with various state-of-the-art algorithms for weakly supervised localization of object in the image like \cite{nguyen2009weakly,hoai2014learning,siva2011weakly,siva2012defence,shi2013bayesian,gokberk2014multi,cinbis2017weakly,wang2014weakly}. We set the parameters of the experiments as follows: (1) Number of proposals per image, $N=1000$. (2) Stopping criteria of iterative spectral clustering, $T=100$. (3) $k$ for NN search, $k=10$. (4) Number of groups to merge, $C=5$. Selecting a higher value of $k$ and $C$ will result into more proposals contributing to the estimation of the final localization window, thus affecting the localization accuracy. We tried with different values of $k$ and $C$ and found that the above values give best results. We use these values for our experiments and keep it constant throughout.

\subsection{Evaluation criteria and runtime}

Following previous works on weakly supervised localization of images, we use the CorLoc (correct location) metric, defined as the percentage of images correctly localized in the whole dataset. Here a correct localization in an image is obtained if the intersection-over-union score of the estimated bounding box and the ground truth bounding box of the image is greater than $0.5$ i.e., $\frac{area(b_{p}\cap b_{gt})}{area(b_{p}\cup b_{gt})}>0.5$. Here $b_{p}$ is the predicted bounding box and $b_{gt}$ is the ground truth bounding box of the same object. This evaluation criteria was suggested by \cite{everingham2010pascal}. Since our algorithm is able to localize a single object instance in an image, we face an issue when the image consists of multiple object instances. This case happens when we evaluate our algorithm on PASCAL VOC 2007 dataset (\cite{pascal-voc-2007}) where many images consists of more than one target object instances. In order to measure CorLoc in such situations, we evaluate our algorithm on per image basis i.e., we classify the image to be correctly localized if any one object instance in the image satisfies the CorLoc condition similar to \cite{wang2014weakly}. We perform the experiments in MATLAB environment on a PC with intel core i7 processor. Our algorithm is able to localize an object in an image of resolution of $500 \times 375$ (of PASCAL VOC dataset) in around $\sim11$ seconds, out of which computing saliency map using \cite{margolin2013makes} technique itself takes $7.5$ secs. In order to make the entire algorithm faster we can replace the saliency algorithm with various other faster alternatives (\cite{cheng2015global}).   
\begin{figure}[h]
\centering
\includegraphics[scale=1.5]{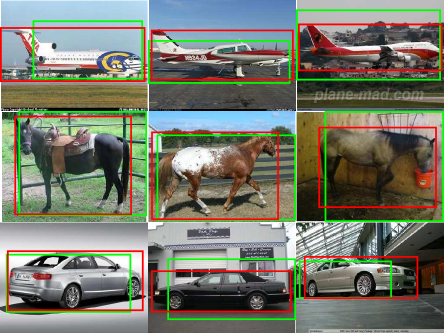}
\caption{Results of localization on Object Discovery dataset (\cite{Rubinstein13Unsupervised}). Green box $=$ Estimated Window, Red box $=$ Ground Truth.}
\label{Fig:results1}
\end{figure}
\subsection{Object Discovery dataset and results}
The Object Discovery dataset (\cite{Rubinstein13Unsupervised}) consists of images from $3$ object categories i.e., aeroplane, car, horse. We evaluate our algorithm on the $100$ image subset. This dataset is mainly used for the purpose of benchmarking object discovery algorithms (\cite{Rubinstein13Unsupervised,cho2015unsupervised,chen_cvpr14}). However, since each image consist of a single object to localize, we can have a much more extensive and better evaluation of our algorithm. No previous weakly supervised techniques for localization with which we benchmark our algorithm (\cite{nguyen2009weakly,hoai2014learning,siva2011weakly,siva2012defence,shi2013bayesian,gokberk2014multi,cinbis2017weakly,wang2014weakly}) have benchmarked their algorithms on this dataset. Since this dataset is used for object discovery algorithms, it consists of noisy images as well. We discard noisy images in the evaluation of our algorithm because of the unavailability of the ground truth for these images. Aeroplane class of the dataset have $18$, car class have $11$ and horse class have $7$ noisy images out of the set of $100$ images each. Thus we get a total of $264$ images for evaluation from the whole dataset. The ground truths are available in the form of segmentations. We convert all the ground truth segmentations into localization boxes. We evaluate our algorithm on all $264$ images of all $3$ classes available in the dataset. The results are shown in Table \ref{tab: ObjDis}. Our algorithm achieves an average CorLoc measure of $51.41 \%$. Images of the results of localization on Object Discovery dataset is shown in Fig. \ref{Fig:results1}. 

\begin{table}[h]
\centering
\caption{Performance of our algorithm on Object Discovery dataset.}
\begin{tabular}{|c|c|c|c|c|}
\hline
\textbf{Class}  & \textbf{Aeroplane} & \textbf{Car} & \textbf{Horse} & \textbf{Average (\%)} \\ \hline
\textbf{CorLoc} & 43.9                & 65.17        & 45.16          & 51.41         \\ \hline
\end{tabular}

\label{tab: ObjDis}
\end{table}

\subsection{PASCAL VOC 2007 dataset and results}
In order to compare our algorithm with various other weakly supervised algorithms, we benchmark it on PASCAL VOC 2007 dataset (\cite{pascal-voc-2007}) which is a very challenging dataset consisting of images captured in real-life scenarios with considerable clutter, occlusion, and diverse viewpoints. It consists of 20 object classes. For a large scale evaluation of the algorithm we take all train+val dataset images which counts to a total of $5011$ images. Each of the $20$ object instances are spread across all the $5011$ images in a range from $215$ (diningtable) to $4690$ (person), having a total of $12608$ object instances. However, as mentioned above we evaluate our algorithm on per image basis i.e., the image is correctly classified in the image if atleast one object instance in the image is correctly localized. 

The results on PASCAL dataset is shown in Table \ref{tab: results}. The second column in the table describes about the amount of data used by these mentioned algorithms. Positive images of the training set is denoted by P and negative images are denoted by N. \cite{wang2014weakly} uses additional data to train a CNN model thus this is represented as A. We evaluate our algorithm on $5011$ images. Our algorithm achieves an average CorLoc measure of about $35.08\%$, which performs better then some of the weakly supervised techniques i.e. \cite{nguyen2009weakly,hoai2014learning,siva2011weakly,siva2012defence} and comparably to \cite{shi2013bayesian}, despite being completely unsupervised. The best performing algorithm is that by \cite{wang2014weakly} which gives an average CorLoc of $48.5\%$. However, it trains a convolutional neural network (CNN) on ILSVRC 2011 dataset in order to extract a $4096$ deep feature descriptor for each proposal of the image. However, we achieve better results than $3$ weakly supervised by just using simple conventional features like HOG and dense SIFT and with the power of spectral clustering. Images of the results of localization on PASCAL VOC dataset is shown in Fig. \ref{Fig:results}.

\begin{table}[h]
\centering
\caption{Performance of our algorithm on PASCAL VOC 2007 dataset.}
\begin{tabular}{@{}ccc@{}}
\toprule
\textbf{Method}           & \textbf{Data used}                                                            & \textbf{CorLoc (\%)} \\ \midrule

\cite{nguyen2009weakly} & P+N                                                                              & 22.4                 \\

\cite{siva2011weakly}   & P+N                                                                              & 30.2                 \\
\cite{siva2012defence}  & P+N                                                                              & 30.4                 \\
\cite{shi2013bayesian}  & P+N                                                                              & 36.2                 \\
\cite{gokberk2014multi} & P+N                                                                              & 38.8                 \\ 

\cite{cinbis2017weakly} & \begin{tabular}[c]{@{}c@{}}P
\end{tabular} & 47.3                 \\

\cite{wang2014weakly}   & P+N+A                                                                       & 48.5                 \\
\bottomrule

\textbf{Ours} &  - & \textbf{35.08}\\
\bottomrule
\end{tabular}

\label{tab: results}
\end{table}
\vspace{-3mm}
\section{Conclusion}
\label{conc}
We have presented a new simple and efficient approach for unsupervised object localization using iterative spectral clustering and proposal grouping. We have shown that our algorithm is able to perform well in challenging scenarios by benchmarking our algorithm on challenging datasets like Object Discovery dataset and PASCAL VOC 2007. We have achieved comparable results in terms of CorLoc when compared to other weakly supervised algorithms. In future work, we plan to extend this work to localizing multiple object instances in an image thus making the algorithm more useful for real-life scenarios.

\bibliographystyle{IEEEtran}
\bibliography{IEEEabrv,ref}
\end{document}